\documentclass{article}
\usepackage[accepted]{icml2017}
\usepackage{times}
\usepackage{amsmath}
\usepackage{amsfonts}
\usepackage{amssymb}
\usepackage{graphicx}
\usepackage[toc,page]{appendix}

\allowdisplaybreaks
\usepackage[breaklinks=true,bookmarks=false]{hyperref}

\icmltitlerunning{Bayesian Belief Updating}

\begin{document} 

\twocolumn[
\icmltitle{Bayesian Belief Updating of Spatiotemporal Dynamics}



\icmlsetsymbol{equal}{*}

\begin{icmlauthorlist}
\icmlauthor{Gerald Cooray}{karo,fil}
\icmlauthor{Richard Rosch}{fil,child}
\icmlauthor{Torsten Baldeweg}{child}
\icmlauthor{Louis Lemieux}{karo}
\icmlauthor{Karl Friston}{fil}
\icmlauthor{Biswa Sengupta}{imp,cam,cor} 
\end{icmlauthorlist}

\icmlaffiliation{karo}{Clinical Neurophysiology, Karolinska University Hospital, Stockholm, Sweden}
\icmlaffiliation{fil}{Wellcome Trust Centre for Neuroimaging, Institute of Neurology, University College London, UK}
\icmlaffiliation{child}{Developmental Neuroscience Programme, Institute of Child Health, University College London, UK}
\icmlaffiliation{karo}{Department of Clinical and Experimental Epilepsy, Institute of Neurology, University College London, UK}
\icmlaffiliation{imp}{Dept. of Bioengineering, Imperial College London}
\icmlaffiliation{cam}{Dept. of Engineering, University of Cambridge}
\icmlaffiliation{cor}{Cortexica Vision Systems, UK}

\icmlcorrespondingauthor{Biswa Sengupta}{b.sengupta@imperial.ac.uk}

\icmlkeywords{Bayesian Belief Updating, Variational Bayes}

\vskip 0.3in
]



\printAffiliationsAndNotice{} 

\begin{abstract} 
Epileptic seizure activity shows complicated dynamics in both space and time. To understand the evolution and propagation of seizures spatially extended sets of data need to be analysed. We have previously described an efficient filtering scheme using variational Laplace that can be used in the Dynamic Causal Modelling (DCM) framework  \cite{Friston2003} to estimate the temporal dynamics of seizures recorded using either invasive or non-invasive electrical recordings (EEG/ECoG). Spatiotemporal dynamics are modelled using a partial differential equation -- in contrast to the ordinary differential equation used in our previous work on temporal estimation of seizure dynamics \cite{Cooray2016}. We provide the requisite theoretical background for the method and test the ensuing scheme on simulated seizure activity data and empirical invasive ECoG data. The method provides a framework to assimilate the spatial and temporal dynamics of seizure activity, an aspect of great physiological and clinical importance. 
\end{abstract} 

\section{Introduction}
 \label{sec:intro}
 
Epilepsy is a chronic disorder characterised by heterogeneous and dynamic pathophysiological processes that lead to an altered balance between excitatory and inhibitory influences at the cortical level. Electrocorticography (ECoG) recordings use a grid of electrodes to cover a cortical area of clinical importance. This methodology allows for a sampling of epileptic seizure activity. Quantitative analysis of the temporal aspect of cortical activity using neural mass (mean-field) models was introduced by Wilson and Cowan \cite{Jansen1995,Wilson1972,Wilson1973}. Variation in the parameters of neural mass models have subsequently been used to model seizure activity, and inferences about these changes can be made using Bayesian statistics; including stochastic filtering or genetic algorithms \cite{Blenkinsop2012,Freestone2014,Nevado-Holgado2012,Schiff2008,Ullah2010,Wendling2005,Sengupta2015,Sengupta2015a}. The inherent assumptions of neural mass models  preclude an in depth analysis of spatiotemporal dynamics, however, other neuronal models, such as neural field models, have explicitly included the spatial aspects of neuronal activity \cite{Moran2013,Pinotsis2012,Schiff2008}. The comparatively high computational complexity of these models makes inference under such models very difficult.

Several studies have suggested that during seizures there are changes in intra- and extracellular factors such as electrolytes, metabolites and neurotransmitters. Mediated through the local interaction between glial and neuronal cells many of these factors will change the intrinsic dynamics of the cortex \cite{Froehlich2010,Ullah2010,Wei2014}. We argue that a simplified spatiotemporal field can model the temporal and spatial behaviour of seizure activity. To contain model complexity and allow inference, we decouple fast temporal activity from slower spatiotemporal fluctuations. This adiabatic assumption respects the biology based on the diffusion or transport of extracellular or intracellular factors as described above and their interaction with the synaptic connectivity of cortical neurons, which are assumed to produce fast temporal activity.

The model is formulated within the Dynamic Causal Modelling (DCM) framework \cite{Friston2003} -- a model comparison and averaging framework for Bayesian inference of hierarchical yet dynamical generative models. This paper introduces a novel application of dynamic causal modelling based upon a hierarchical DCM, with first (fast) level spectral activity and second (slow) level spatial dynamics.  

\section{Methods}
 \label{sec:methods}
 
To construct an invertible generative model, we partition our model into hidden states/parameters that are not explicitly (conditionally) dependent on spatial location $\left( x, \theta _{i}, \theta _{e} \right)$ and parameters that depend on spatial location $\left(\theta _{sp} \right)$. Furthermore, we assume that the temporal fluctuations of the spatially dependent subset will be several orders slower than the temporal dynamics of the hidden states/parameters that are conditionally independent of spatial location. Parameters not governed by spatial dynamics are sampled from a Gaussian distribution. For reasons of numerical efficiency, we introduce the simplifying assumption that transport of electrolytes through the extracellular medium or through glial cells is via passive diffusion.

At this point we appeal to the assumption that the dynamics of the neuronal activity ($x$) compared to the spatially varying parameters ($\theta _{sp}$) are at least several orders of magnitude faster. This allows us to estimate neuronal activity as the steady state activity of a neural mass governing $\textit{x}$. We can then formulate the expected spectrum of the neuronal activity as a function of the parameters $\left(  \theta _{i}, \theta _{sp}, \theta _{e} \right)$. This approach has been adopted in a similar setting \cite{Cooray2016,Moran2011}.

Our model of the spectra measured during discrete time intervals from the local field potentials is given by the following measurement model and a partial differential equation (Eq. \ref{eqn:eqn1_2}): 

\begin{eqnarray}
y_{k} & = & H \left(  \theta _{i} \left( t_{k} \right) , \theta _{sp} \left( t_{k} \right) ,  \theta _{e} \right) + r_{k} \nonumber \\
\frac{d \theta _{sp}}{dt} & = & \triangledown ^{2} \theta _{sp}+q \left( t \right) 
 \label{eqn:eqn1_2}
\end{eqnarray}

In the following equations, sampling error $ r_{k} \sim \mathcal{N} \left( 0,R_{k} \right)$ is sampled from a Gaussian distribution and $ q \left( t \right)$ is defined as a white noise process (Eq. \ref{eqn:eqn3}). The diffusion equation is infinite dimensional, making it difficult to solve analytically and numerically. However, we approximate the partial differential equation using a finite set of eigenfunctions, $\phi _{i} \left( x \right)$, and eigenvalues, $\lambda _{i} $. This  results in the following  set of equations \cite{Saerkkae2012,Solin2012}:
 
\begin{eqnarray}
y_{k} & = & H \left(  \theta _{i} \left( x,t_{k} \right) , \theta _{sp} \left( x,t_{k} \right) , \theta _{e} \left( x,t_{k} \right)  \right) +r_{k} \nonumber \\
{\theta _{sp}}\left( t \right) & \approx & \sum {{c_i}\left( {{t_k}} \right){\phi _i}\left( x \right){e^{ - {\lambda _i}\left( {t - {t_k}} \right)}}} \nonumber \\
 c_{i} \left( t_{k} \right) & = & c_{i} \left( t_{k-1} \right) e^{- \lambda _{i} \left( t_{k}-t_{k-1} \right) }+q_{i} \left( t_{k-1} \right) 
 \label{eqn:eqn3}
\end{eqnarray}

In this equation $q_{i} \left( t_{k} \right)  \sim \mathcal{N} \left( 0,Q_{i,k} \right) $ is sampled from a Gaussian distribution. A similar idea has previously been presented in DCM of distributed electromagnetic responses where the cortical activity was parameterised using a set of local standing-waves of neural field models \cite{Daunizeau2009}. 

The data generated by this model comprised of windowed spectral data ${y_{k,ij}}$  from the $i^{th}$ row and the $j^{th}$ column of any electrode array used to measure the electrographic activity sampled at time $t_{k}$. We estimate the parameters of our model in each time window using variational Laplace \cite{MacKay2002}. The values in the next time window are predicted using Eq. \ref{eqn:eqn3}, given the previous values. We can then estimate an approximate solution to the above stochastic model using  Bayesian belief updating as has been previously described \cite{Cooray2016}.  For technical details please refer to Section \ref{appendix:canonical}-\ref{appendix:neuralfield}.

\section{Results}
 \label{sec:results}
 
\begin{figure}
\includegraphics[trim={5cm 4cm 5cm 5cm},clip,angle=270,scale=0.5]{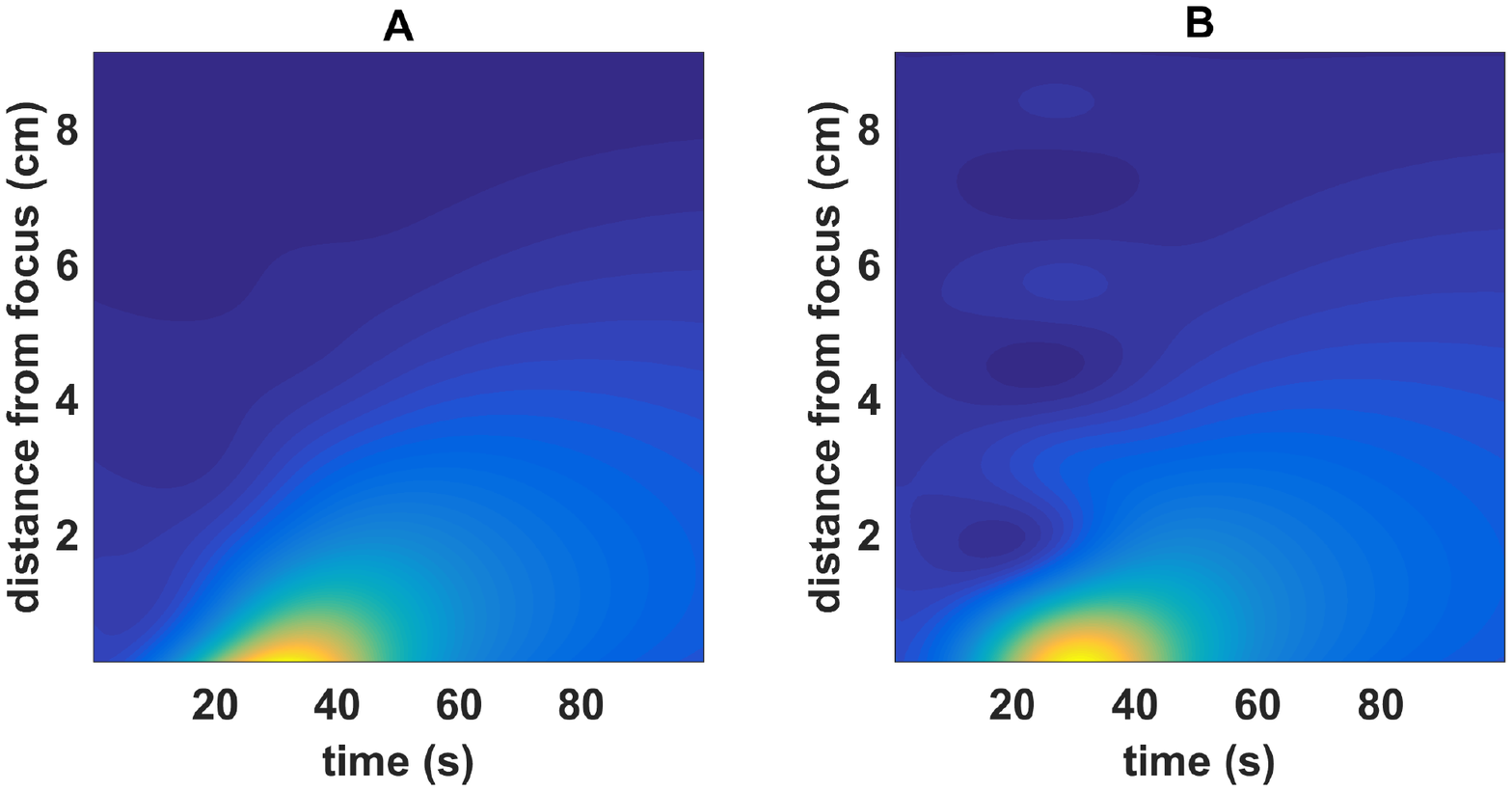}
\caption{Left pane: Simulated fluctuation in parameter. Right pane: Results of inversion of simulated data (not shown).}
\label{fig:fig1}
\end{figure}

We simulated ECoG data with two-dimensional spatial dynamics: giving an excellent fit to the spectral activity (explained variation $>$ 0.99; Figure \ref{fig:fig1}). Furthermore, the inferred fluctuations of the excitatory parameter were nearly identical to the simulated dynamics.

\begin{figure}
\includegraphics[scale=0.4,trim={5cm 6cm 4cm 6cm},clip]{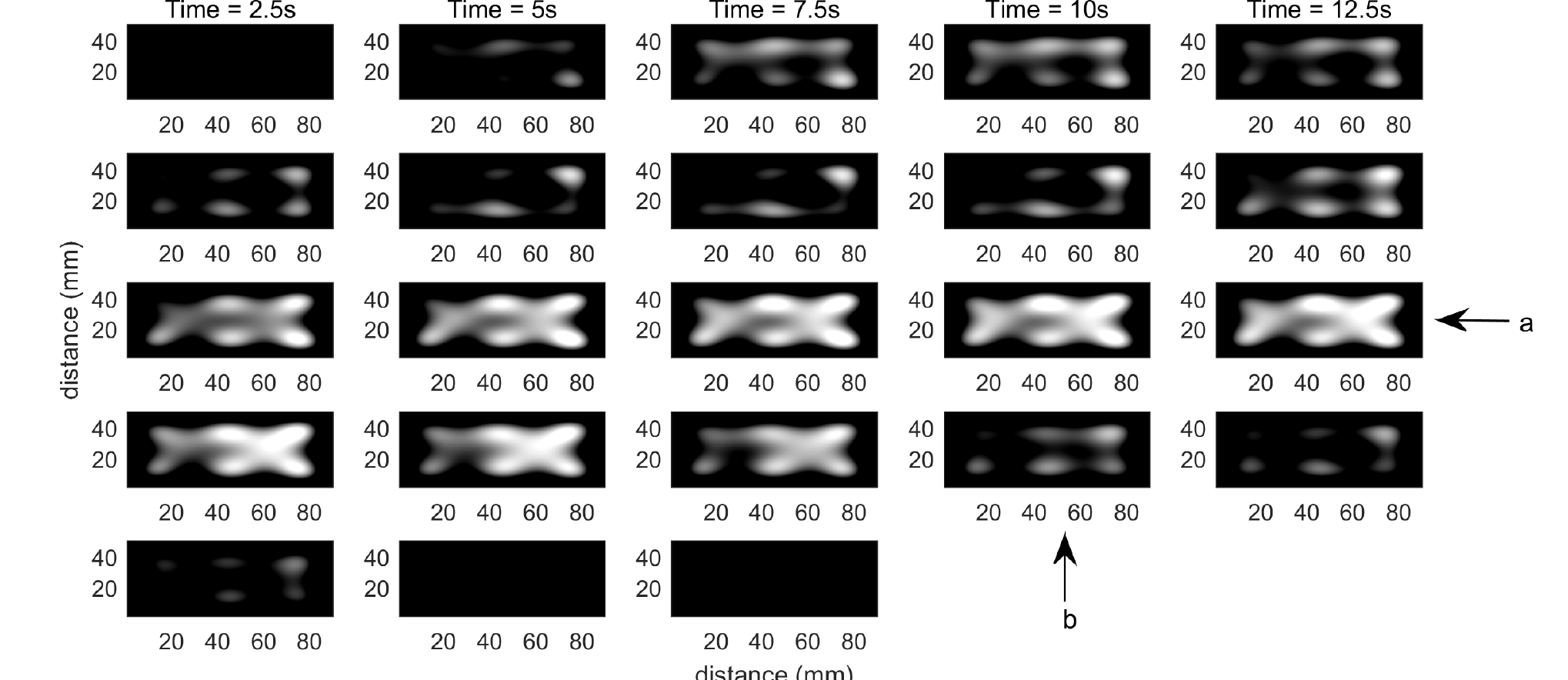}
\caption{Estimated cortical excitability at different time points}
 \label{fig:fig2}
\end{figure}

We also estimated a spatially dependent excitatory gain parameter for epileptic seizure activity using recordings from two patients with focal cortical dysplasia and refractory epilepsy (Figure \ref{fig:fig2}). Subdural recordings were made from a 32-contact and a 20-contact platinum array (Ad-Tech) placed over the lesion. We used two seizures free of artefacts from each data set for modelling the averaged induced spectral activity. The model also estimated parameters (without spatiotemporal dynamics) of intrinsic connectivity within the cortical column that were sampled from a white noise process. We allowed this random variation during Bayesian belief updating to accommodate unmodelled changes during the seizure activity. There was a good fit between the data and the predicted activity (explained variation $>$ 0.85).

Spatial dynamics controlling cortical excitability were inferred from the data, enabling parameters (excitatory gain) to be estimated at each point within the cortical grid over time. The parameter increased to a maximum towards the end of the seizure (a) and then quickly dissipates during termination of the seizure (b) (Figure \ref{fig:fig2}).

\section{Discussion}
 \label{sec:discuss}
 
We present a framework for analysing spatiotemporal dynamics, in our specific case such a dynamics emerges from epileptic seizure activity. We extended an inference scheme previously described for inference of the temporal structure of seizure activity \cite{Cooray2016,Cooray2017} -- we achieved this objective using a field model to incorporate spatial dynamics. Including spatial processes allow us to model the epileptic spread of seizure activity, which is one of the key physiological processes underlying a focal seizure. A deeper understanding of the spatial spread of focal seizures over the cortex could be used for curtailing the spread of a seizure; limiting its effect on the patient by either surgical or other procedures such as direct or induced electrical currents delivered by e.g. transcranial magnetic stimulation or deep brain stimulation  \cite{Engel1993,Rotenberg2008}.

The current framework for the analysis of seizure activity assumes that the spatial dynamics can be estimated using a finite set of eigenfunctions of the underlying partial differential equation (modelling the spatial dynamics). This assumption is only valid for relatively simple spatiotemporal dynamics (partial differential equations with analytical solutions) on simple representations of the cortical surface (such as a plane or spherical surface) \cite{Evans2010}. If any of these assumptions are violated, as in the case of realistic geometries obtained using MRI or more realistic descriptions of the spatiotemporal dynamics, the forward model would require re-formulation using numerical approximations. One way forward is by reformulating our scheme under a finite element framework using the approach described in \citet{Sengupta2017}; although this requires extensive computational expenditure. The real value of this work, therefore, lies in the potential to fit the parameters of spatiotemporal dynamics of generative models to individual patients. This should allow us to understand individual differences that would not only guide epileptogenic resections but also be used for prediction of epilepsy surgery outcomes. 

In conclusion, we present a straightforward way of extending an established Bayesian belief update/filtering scheme to include the spatial spread of seizure activity. This requires several (plausible) assumptions regarding the behaviour of cortical columns, the geometry of the cortex and a separation of time scales of fast neuronal activity and slower fluctuations in cortical gain. We suggest that these assumptions have some experimental evidence but that the method can be modified if these assumptions are shown to be invalid or too inaccurate in the future. In particular, by comparing spatiotemporal models of different complexity (e.g. diffusion models versus complex reaction-diffusion models with threshold dynamics), using their model evidence, the above framework could be used to address important questions about how seizure activity spreads in patients. 

{\small
\bibliography{spatio}

\begin{thebibliography}{27}
\providecommand{\natexlab}[1]{#1}
\providecommand{\url}[1]{\texttt{#1}}
\expandafter\ifx\csname urlstyle\endcsname\relax
  \providecommand{\doi}[1]{doi: #1}\else
  \providecommand{\doi}{doi: \begingroup \urlstyle{rm}\Url}\fi

\bibitem[Blenkinsop et~al.(2012)Blenkinsop, Valentin, Richardson, and
  Terry]{Blenkinsop2012}
Blenkinsop, Alex, Valentin, Antonio, Richardson, Mark~P, and Terry, John~R.
\newblock The dynamic evolution of focal-onset epilepsies--combining
  theoretical and clinical observations.
\newblock \emph{European Journal of Neuroscience}, 36\penalty0 (2):\penalty0
  2188--2200, 2012.

\bibitem[Cooray et~al.(2016)Cooray, Sengupta, Douglas, and Friston]{Cooray2016}
Cooray, Gerald~K, Sengupta, Biswa, Douglas, Pamela~K, and Friston, Karl.
\newblock Dynamic causal modelling of electrographic seizure activity using
  {Bayesian} belief updating.
\newblock \emph{NeuroImage}, 125:\penalty0 1142--1154, 2016.

\bibitem[{Cooray} et~al.(2017){Cooray}, {Rosch}, {Baldeweg}, {Lemieux},
  {Friston}, and {Sengupta}]{Cooray2017}
{Cooray}, G.K, {Rosch}, R., {Baldeweg}, T., {Lemieux}, L., {Friston}, K., and
  {Sengupta}, B.
\newblock {Bayesian Belief Updating of Spatiotemporal Seizure Dynamics}.
\newblock \emph{ArXiv e-prints}, 2017.

\bibitem[Daunizeau et~al.(2009)Daunizeau, Kiebel, and Friston]{Daunizeau2009}
Daunizeau, Jean, Kiebel, Stefan~J, and Friston, Karl~J.
\newblock Dynamic causal modelling of distributed electromagnetic responses.
\newblock \emph{NeuroImage}, 47\penalty0 (2):\penalty0 590--601, 2009.

\bibitem[Engel(1993)]{Engel1993}
Engel, J.
\newblock \emph{Surgical treatment of the epilepsies}.
\newblock Raven Press, 1993.

\bibitem[Evans(2010)]{Evans2010}
Evans, CE.
\newblock \emph{Partial Differential Equations}.
\newblock American Mathematical Society, 2010.

\bibitem[Freestone et~al.(2014)Freestone, Karoly, Ne{\v{s}}i{\'c}, Aram, Cook,
  and Grayden]{Freestone2014}
Freestone, Dean~R, Karoly, Philippa~J, Ne{\v{s}}i{\'c}, Dragan, Aram, Parham,
  Cook, Mark~J, and Grayden, David~B.
\newblock Estimation of effective connectivity via data-driven neural modeling.
\newblock \emph{Frontiers in neuroscience}, 8, 2014.

\bibitem[Friston et~al.(2003)Friston, Harrison, and Penny]{Friston2003}
Friston, Karl~J, Harrison, Lee, and Penny, Will.
\newblock Dynamic causal modelling.
\newblock \emph{Neuroimage}, 19\penalty0 (4):\penalty0 1273--1302, 2003.

\bibitem[Fr{\"o}hlich et~al.(2010)Fr{\"o}hlich, Sejnowski, and
  Bazhenov]{Froehlich2010}
Fr{\"o}hlich, Flavio, Sejnowski, Terrence~J, and Bazhenov, Maxim.
\newblock Network bistability mediates spontaneous transitions between normal
  and pathological brain states.
\newblock \emph{Journal of Neuroscience}, 30\penalty0 (32):\penalty0
  10734--10743, 2010.

\bibitem[Jansen \& Rit(1995)Jansen and Rit]{Jansen1995}
Jansen, Ben~H and Rit, Vincent~G.
\newblock Electroencephalogram and visual evoked potential generation in a
  mathematical model of coupled cortical columns.
\newblock \emph{Biological cybernetics}, 73\penalty0 (4):\penalty0 357--366,
  1995.

\bibitem[MacKay(2002)]{MacKay2002}
MacKay, David J.~C.
\newblock \emph{Information Theory, Inference \& Learning Algorithms}.
\newblock Cambridge University Press, New York, NY, USA, 2002.

\bibitem[Moran et~al.(2013)Moran, Pinotsis, and Friston]{Moran2013}
Moran, Rosalyn, Pinotsis, Dimitris~A, and Friston, Karl.
\newblock Neural masses and fields in dynamic causal modeling.
\newblock \emph{Frontiers in computational neuroscience}, 7, 2013.

\bibitem[Moran et~al.(2011)Moran, Stephan, Dolan, and Friston]{Moran2011}
Moran, Rosalyn~J, Stephan, Klaas~E, Dolan, Raymond~J, and Friston, Karl~J.
\newblock Consistent spectral predictors for dynamic causal models of
  steady-state responses.
\newblock \emph{Neuroimage}, 55\penalty0 (4):\penalty0 1694--1708, 2011.

\bibitem[Nevado-Holgado et~al.(2012)Nevado-Holgado, Marten, Richardson, and
  Terry]{Nevado-Holgado2012}
Nevado-Holgado, Alejo~J, Marten, Frank, Richardson, Mark~P, and Terry, John~R.
\newblock Characterising the dynamics of {EEG} waveforms as the path through
  parameter space of a neural mass model: application to epilepsy seizure
  evolution.
\newblock \emph{Neuroimage}, 59\penalty0 (3):\penalty0 2374--2392, 2012.

\bibitem[Pinotsis et~al.(2012)Pinotsis, Moran, and Friston]{Pinotsis2012}
Pinotsis, Dimitris~A, Moran, Rosalyn~J, and Friston, Karl~J.
\newblock Dynamic causal modeling with neural fields.
\newblock \emph{Neuroimage}, 59\penalty0 (2):\penalty0 1261--1274, 2012.

\bibitem[Rotenberg et~al.(2008)Rotenberg, Muller, Birnbaum, Harrington,
  Riviello, Pascual-Leone, and Jensen]{Rotenberg2008}
Rotenberg, Alexander, Muller, Paul, Birnbaum, Daniel, Harrington, Michael,
  Riviello, James~J, Pascual-Leone, Alvaro, and Jensen, Frances~E.
\newblock Seizure suppression by {EEG}-guided repetitive transcranial magnetic
  stimulation in the rat.
\newblock \emph{Clinical Neurophysiology}, 119\penalty0 (12):\penalty0
  2697--2702, 2008.

\bibitem[S{\"a}rkk{\"a} et~al.(2012)S{\"a}rkk{\"a}, Solin, Nummenmaa, Vehtari,
  Auranen, Vanni, and Lin]{Saerkkae2012}
S{\"a}rkk{\"a}, Simo, Solin, Arno, Nummenmaa, Aapo, Vehtari, Aki, Auranen,
  Toni, Vanni, Simo, and Lin, Fa-Hsuan.
\newblock Dynamic retrospective filtering of physiological noise in {BOLD fMRI:
  DRIFTER}.
\newblock \emph{NeuroImage}, 60\penalty0 (2):\penalty0 1517--1527, 2012.

\bibitem[Schiff \& Sauer(2008)Schiff and Sauer]{Schiff2008}
Schiff, Steven~J and Sauer, Tim.
\newblock Kalman filter control of a model of spatiotemporal cortical dynamics.
\newblock \emph{BMC Neuroscience}, 9\penalty0 (1):\penalty0 O1, 2008.

\bibitem[Sengupta et~al.(2015{\natexlab{a}})Sengupta, Friston, and
  Penny]{Sengupta2015}
Sengupta, B., Friston, K.~J., and Penny, W.~D.
\newblock Gradient-based {MCMC} samplers for dynamic causal modelling.
\newblock \emph{Neuroimage}, 2015{\natexlab{a}}.

\bibitem[Sengupta et~al.(2015{\natexlab{b}})Sengupta, Friston, and
  Penny]{Sengupta2015a}
Sengupta, B., Friston, K.~J., and Penny, W.~D.
\newblock Gradient-free {MCMC} methods for dynamic causal modelling.
\newblock \emph{Neuroimage}, 112:\penalty0 375--81, 2015{\natexlab{b}}.

\bibitem[Sengupta \& Friston(2017)Sengupta and Friston]{Sengupta2017}
Sengupta, Biswa and Friston, Karl.
\newblock Sentient self-organization: Minimal dynamics and circular causality.
\newblock \emph{arXiv preprint arXiv:1705.08265}, 2017.

\bibitem[Solin et~al.(2012)]{Solin2012}
Solin, Arno et~al.
\newblock Hilbert space methods in infinite-dimensional {Kalman} filtering.
\newblock 2012.

\bibitem[Ullah \& Schiff(2010)Ullah and Schiff]{Ullah2010}
Ullah, Ghanim and Schiff, Steven~J.
\newblock Assimilating seizure dynamics.
\newblock \emph{PLoS computational biology}, 6\penalty0 (5):\penalty0 e1000776,
  2010.

\bibitem[Wei et~al.(2014)Wei, Ullah, Ingram, and Schiff]{Wei2014}
Wei, Yina, Ullah, Ghanim, Ingram, Justin, and Schiff, Steven~J.
\newblock Oxygen and seizure dynamics: {II. Computational modeling}.
\newblock \emph{Journal of neurophysiology}, 112\penalty0 (2):\penalty0
  213--223, 2014.

\bibitem[Wendling et~al.(2005)Wendling, Hernandez, Bellanger, Chauvel, and
  Bartolomei]{Wendling2005}
Wendling, Fabrice, Hernandez, Alfredo, Bellanger, Jean-Jacques, Chauvel,
  Patrick, and Bartolomei, Fabrice.
\newblock Interictal to ictal transition in human temporal lobe epilepsy:
  insights from a computational model of intracerebral {EEG}.
\newblock \emph{Journal of Clinical Neurophysiology}, 22\penalty0 (5):\penalty0
  343, 2005.

\bibitem[Wilson \& Cowan(1972)Wilson and Cowan]{Wilson1972}
Wilson, Hugh~R and Cowan, Jack~D.
\newblock Excitatory and inhibitory interactions in localized populations of
  model neurons.
\newblock \emph{Biophysical journal}, 12\penalty0 (1):\penalty0 1--24, 1972.

\bibitem[Wilson \& Cowan(1973)Wilson and Cowan]{Wilson1973}
Wilson, Hugh~R and Cowan, Jack~D.
\newblock A mathematical theory of the functional dynamics of cortical and
  thalamic nervous tissue.
\newblock \emph{Biological Cybernetics}, 13\penalty0 (2):\penalty0 55--80,
  1973.

\end{thebibliography}
\bibliographystyle{icml2017}
}

\onecolumn

\begin{appendices}

\section{Canonical mean-field equations}
\label{appendix:canonical}

The cortical columns described in this note were modelled using a neural mass (mean-field) model generating electrographic seizure activity. The canonical cortical microcircuit (CMC) is comprised of four subpopulations of neurons corresponding to superficial and deep pyramidal, excitatory, and inhibitory cells \cite{Moran2013}. These cell populations are connected using ten inhibitory and excitatory connections. Afferent connections drive the excitatory granular cells and efferent connections derive from the superficial pyramidal cells. The equations governing the four mean-fields are given by,

\begin{eqnarray}
{\ddot x_e} + \frac{{2{{\dot x}_e}}}{{{T_e}}} + \frac{{{x_e}}}{{T_e^2}}{\rm{ }} & = & {\rm{ }} - {g_1}s\left( {{x_e}} \right) - {g_3}s\left( {{x_i}} \right) - {g_2}s\left( {{x_{sp}}} \right) \nonumber \\
{\ddot x_i} + \frac{{2{{\dot x}_i}}}{{{T_i}}} + \frac{{{x_i}}}{{T_i^2}}{\rm{ }} & = & {\rm{ }}{\theta _{sp}}{g_5}s\left( {{x_e}} \right) + {\theta _{sp}}{g_6}s\left( {{x_{dp}}} \right) - {g_4}s\left( {{x_i}} \right) \nonumber \\
{\ddot x_{sp}} + \frac{{2{{\dot x}_{sp}}}}{{{T_{sp}}}} + \frac{{{x_{sp}}}}{{T_{sp}^2}}{\rm{ }} & = & {\rm{ }}{\theta _{sp}}{g_8}s\left( {{x_e}} \right) - {g_7}s\left( {{x_{sp}}} \right) \nonumber \\
{\ddot x_{dp}} + \frac{{2{{\dot x}_{dp}}}}{{{T_{dp}}}} + \frac{{{x_{dp}}}}{{T_{dp}^2}}{\rm{ }} & = & {\rm{ }} - {g_{10}}s\left( {{x_{dp}}} \right) - {g_9}s\left( {{x_i}} \right)\nonumber \\
\end{eqnarray}

 $\theta_{sp}$ is the parameter with spatiotemporal dynamics. Parameter $g_7$ was sampled from a Gaussian distribution during  Bayesian filtering. All other parameters were kept constant during the inversion. 

\section{Solution of parameter field equations}
\label{appendix:field}
 
The spatio-temporal (on a single spatial dimension) parameter is modelled as a heat equation with time dependent boundary conditions,

\begin{eqnarray}
u \left( x,t \right)  & : & \text{parameter with spatiotemporal variation} \nonumber \\
\alpha  & : & \text{diffusion coefficient} \nonumber \\
\phi _{1/0} \left( t \right)  & : & \text{boundary condition (time dependent)} \nonumber \\
f \left( x \right)  & : & \text{initial condition} \nonumber \\
L  & : & \text{length of one dimensional domain over which \textit{u} varies} \nonumber \\
\omega  \left( x \right)  & : & \text{time independent particular solution} \nonumber \\
v \left( x,t \right)  & : & \text{homogenous solution to diffusion equation} \nonumber \\
S \left( x,t \right)  & : & \text{auxillary variable} \nonumber \\
 \nonumber \\
{u_t} & = & {\alpha ^2}{u_{xx}} \nonumber \\
u\left( {0,t} \right){\rm{ }} & = & {\rm{ }}{\phi _0},\ u\left( {L,t} \right) = {\phi _1} \nonumber \\
u\left( {x,0} \right){\rm{ }} & = & {\rm{ }}f\left( x \right)  \nonumber \\
\omega {\rm{ }} & = & {\rm{ }}{\phi _0} + \frac{x}{L}\left( {{\phi _1} - {\phi _0}} \right)   \nonumber \\
u{\rm{ }} & = & {\rm{ }}\omega  + v \nonumber \\
{u_t} & = & {\omega _t} + {v_t} = {\alpha ^2}{v_{xx}}\nonumber \\
{v_t} & = & {\alpha ^2}{v_{xx}} - {\omega _t}\nonumber \\
v\left( {0,t} \right){\rm{ }} & = & {\rm{ }}v\left( {L,t} \right) = 0 \nonumber \\
v\left( {x,0} \right){\rm{ }} & = & {\rm{ }}f\left( x \right) - \omega \left( {x,0} \right) \nonumber \\
S\left( {x,t} \right){\rm{  }} & = & {\rm{ }} - \dot \omega 
\end{eqnarray}

The solution to the diffusion equation is separated into time and space dependent functions; i.e., by using separation of variables. The full solution can then be written as an infinite series of eigenfunctions (space dependent functions), together with their time-dependent variation. The eigenfunctions of the diffusion equation satisfy an equivalent ordinary differential equation.  For the one-dimensional diffusion equation with Dirichlet boundary conditions (zero along boundaries), the eigenfunctions can be written as sine functions. For the two-dimensional case with rotational symmetry, a linear combination of first and second order Bessel functions are used,

\begin{eqnarray}
\lambda _{n} & : & \text{eigenvalue of ordinary differential equation}   \nonumber \\
{S_n}\left( t \right) & : & \text{coefficient of the eigenfunction } sin{\lambda _n}x  
\text{ in the eigenfunction expansion of the auxiliary variable }  S\left( {x,t} \right)   \nonumber \\
v_{n} \left( t \right)  & : & \text{coefficient of the eigenfunction }  sin \lambda _{n}x \text{ in the eigenfunction expansion of } v \left( x,t \right)  \nonumber \\
c_{n} & : & \text{coefficient of the eigenfunction }  sin \lambda _{n}x \text{ in the eigenfunction expansion of the initial condition of }   v \left( x,t \right), \text{ i.e., }  v \left( x,0 \right)   \nonumber \\
 \nonumber \\
S \left( x,t \right) & = & \sum _{n=1}^{ \infty}S_{n} \left( t \right) sin \lambda _{n}x\nonumber \\
S_{n} \left( t \right) & = & -\frac{2}{L} \int _{0}^{L}\frac{d \omega }{dt} sin \lambda _{n}xdx\nonumber \\
\lambda _{n}& = & \frac{n \pi }{L} \nonumber \\
v \left( x,t \right) & = & \sum _{n=1}^{ \infty}v_{n} \left( t \right) sin \lambda _{n}x \nonumber \\
v \left( x,0 \right) & = & \sum _{n=1}^{ \infty}c_{n}sin \lambda _{n}x \nonumber \\
c_{n} & = & \frac{2}{L} \int _{0}^{L} \left( f \left( x \right) - \omega  \left( x,0 \right)  \right) sin \lambda _{n}xdx \nonumber \\
u\left( {x,t} \right){\rm{ }} & = & {\rm{ }}\omega \left( {x,t} \right) + \sum\limits_{n = 1}^\infty  {\left( {\int_0^t {{S_n}} \left( t \right)exp\left( {{\alpha ^2}\lambda _n^2\tau } \right)d\tau  + {c_n}} \right)} {\rm{ }}exp\left( { - {\alpha ^2}\lambda _n^2t} \right)sin{\lambda _n}x\nonumber \\
\end{eqnarray}

\section{Bayesian belief updating}
\label{appendix:bayes}

The following equations operationalize the Bayesian belief updating of the parameters as data is inverted sequentially across the windowed data, $y_{i}$, see \cite{Cooray2016,Cooray2017} for more details. The priors on the $n$ eigenfunction coefficients of parameter $\theta _{sp}$ in each window is given by,

\begin{eqnarray}
\mu _{i} & : & \text{posterior mean vector of }  \theta _{sp}  \text{ in the ith window i.e., }   \left( c_{1},c_{2},\ldots,c_{n} \right) _{i} \nonumber \\
\lambda  & : & \text{eigenvalues, i.e. }   \left(  \lambda _{1}, \lambda _{2},\ldots, \lambda _{n} \right)  \nonumber \\
\Delta  & : & \text{time interval between windows} \nonumber \\
Q_{i}  & : & \text{posterior covariance matrix of }  \theta _{sp} \text{ in the ith window} \nonumber \\
R & : & \text{volatility covariance matrix of }   \theta _{sp} \nonumber \\
p \left(  \theta _{sp} \vert y_{i},\ldots,y_{1} \right)  & = & \text{prior probability density of }   \theta _{sp} \text{ in the i+1th  window} \nonumber \\
p \left(  \theta _{sp} \vert y_{i},\ldots,y_{1} \right) & = & \mathcal{N} \left(  \mu _{i}e^{- \lambda  \Delta },Q_{i}+R \right)  \nonumber \\
\end{eqnarray}

Using a variational Bayes formulation, we update this prior probability distribution using the data in window $i+1$.

\section{Parameterizations of field models}
\label{appendix:neuralfield}

\subsection*{Neural field}

\begin{eqnarray}
u \left( x,t \right)  & : & \text{neural field representing activity of a population of neurons at location x at time t}\nonumber \\
w \left( y \right)  & : & \text{the strength of connections between neurons separated by a distance y} \nonumber \\ 
f \left( u \right)   & : & \text{firing rate function,} \nonumber \\
u_{t} \left( x,t \right) & = & -u+  \int _{- \infty}^{ \infty} w \left( y \right) f \left( u \left( x-y,t \right)  \right)  dy\nonumber \\
\end{eqnarray}

\subsection*{Parameter field}

Note that the spatial variation of the \textit{neural field} is due to the time independent variable $w$. As this term is convolved over space with the firing rate it becomes difficult to separate the spatial and the temporal dynamics. In the case of the \textit{parameter field} it is possible to have a slowly diffusing parameter (diffusion rate dependent on $\alpha$) that interacts as though it is constant with respect to the dynamics of the faster neural mean fields (rate dependent on $k$).

\begin{eqnarray}
u \left( x,t \right)  & : & \text{neural field representing activity of a population of neurons at location x at time t} \nonumber \\
f \left( u \right)  & :  & \text{firing rate function}  \nonumber \\
\theta  \left( x,t \right)  & :  & \text{parameter field} \nonumber \\
k & :  & \text{rate coefficient of neural mass} \nonumber \\
\alpha  & :  & \text{diffusion coefficient} \nonumber \\
\nonumber \\
{u_{tt}}\left( {x,t} \right) - \;2k{u_t}\left( {x,t} \right) - {k^2}u\left( {x,t} \right){\rm{ }} & = & {\rm{ }}f\left( {u\left( {x,t} \right),\;\theta \left( {x,t} \right)} \right)\nonumber \\
{\theta _t}{\rm{ }} & = & {\rm{ }}{\alpha ^2}{\theta _{xx}} \nonumber \\
\end{eqnarray}

\end{appendices}


\end{document}